\pdfoutput=1
\documentclass[letterpaper, 10pt, conference]{ieeeconf}

\IEEEoverridecommandlockouts   %
\usepackage{graphicx}
\usepackage{amsmath,amssymb,bm}
\usepackage{booktabs}
\usepackage{multirow}
\usepackage[table]{xcolor}     %
\usepackage{array}
\usepackage{url}
\usepackage[draft,bookmarks=false]{hyperref}

\makeatletter
\let\streamrig@makecaption\@makecaption
\long\def\@makecaption#1#2{%
  \ifx\@captype\@IEEEtablestring
    \begin{center}{\footnotesize #1.\enspace{\scshape #2}}\end{center}%
    \vskip 2pt\relax %
  \else
    \streamrig@makecaption{#1}{#2}%
  \fi}
\makeatother

\newcommand{\ours}{StreamRig}                  %
\newcommand{\prior}{G2G}                       %

\newcommand{\SE}{\mathrm{SE}(3)}

\providecommand{\R}{}\renewcommand{\R}{\mathbb{R}}
\newcommand{\pose}[2]{T_{#1 \leftarrow #2}}    %
\definecolor{cBest}{HTML}{0091AD}
\definecolor{cSecond}{HTML}{D2791E}
\newcommand{\best}[1]{\textcolor{cBest}{#1}}
\newcommand{\second}[1]{\textcolor{cSecond}{#1}}

\title{\LARGE \bf
StreamRig: Exploiting Intra-Rig Geometry for Streaming Multi-Camera Odometry}

\author{Yufei Wei$^{1}$, Shuhao Ye$^{1}$, Qi Wang$^{2}$, Xin Zheng$^{1}$, Qing Huang$^{1}$, Rong Xiong$^{1}$, and Yue Wang$^{1,*}$}

\begin{document}

\maketitle
\thispagestyle{empty}
\pagestyle{empty}

\begin{abstract}
Mobile robots and vehicles carry synchronized multi-camera rigs, yet many streaming 3D foundation models are designed for monocular input, leaving efficient use of rig geometry a challenge. We present \ours{}, a \emph{freeze-and-stream} framework that builds causal streaming odometry for calibrated rigs on a frozen multi-view 3D foundation model. The frozen front-end jointly perceives the synchronized views using rig calibration. A \emph{Rig-Resampler} compresses their features, a \emph{CausalBridge} applies causal attention with a key-value cache, and a lightweight head regresses rig poses. A periodic re-anchoring protocol supports stable pose estimation over long sequences. Only these modules are trained, 74.6\,M parameters in total, with relative poses as the sole supervision. Our two-stage training strategy combines group relocalization pretraining with causal rig training to transfer the geometric priors of the frozen front-end and the alignment ability of the pretrained modules to streaming odometry. We evaluate on NCLT, TartanGround, KITTI-360, and our self-collected humanoid-robot dataset ZJH, where training uses only simulation and real-world evaluation is zero-shot. Across all four datasets, \ours{} achieves lower translation and rotation drift than the evaluated non-oracle monocular streaming and rig-aware offline models, while maintaining low inference cost. Ablations and controlled camera-count experiments identify the sources of these gains. We further examine how longer training windows affect inference over longer horizons.
Code has been released at \url{https://github.com/WeiYuFei0217/StreamRig}.
\end{abstract}

\vspace{3pt} %

\begin{figure}[t]
  \centering
  \includegraphics[width=0.9\columnwidth]{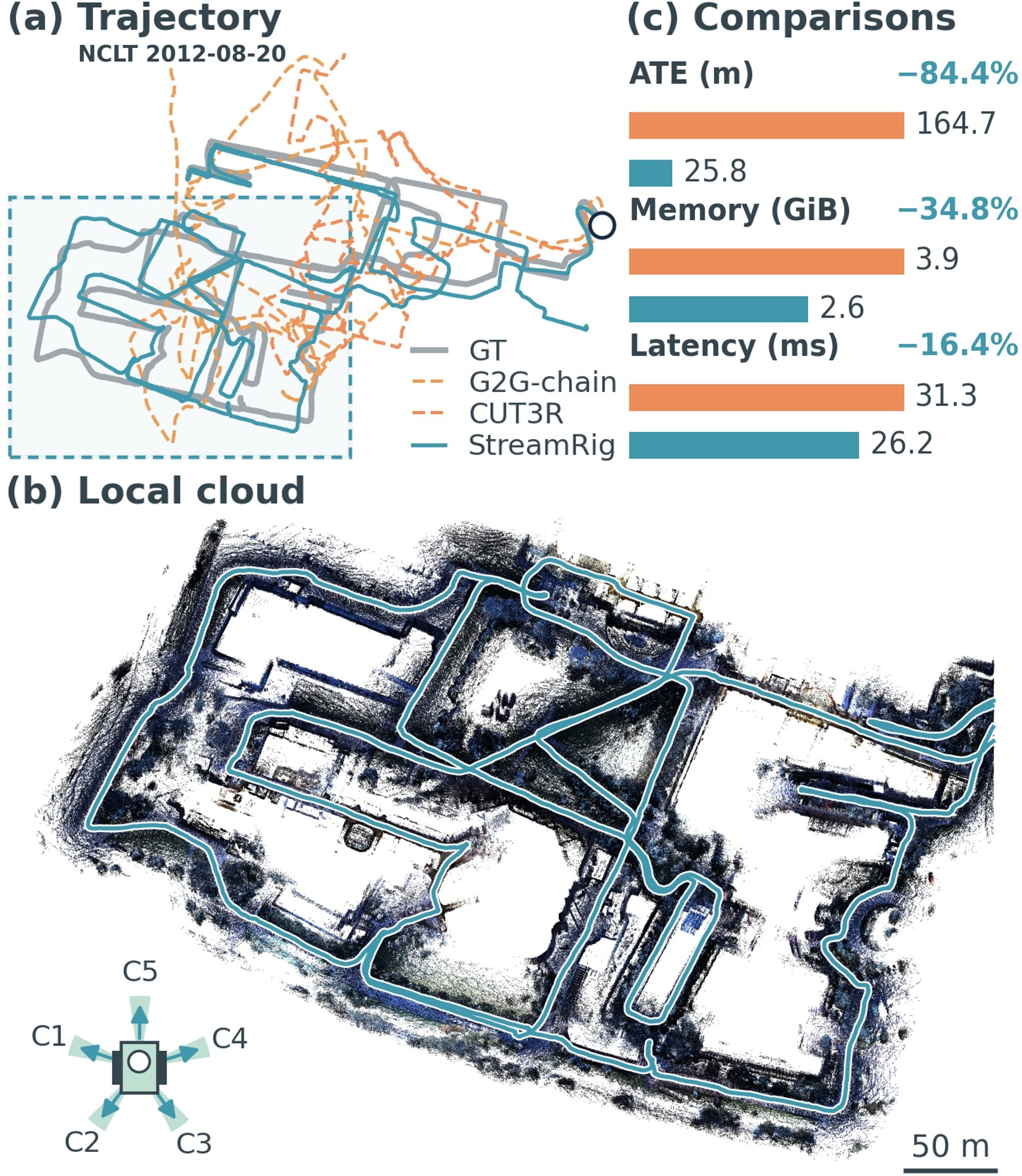}
  \vspace{-3pt} %
  \caption{\textbf{Accurate and efficient surround-view odometry.}
  (a) \ours{} follows the NCLT ground-truth trajectory more closely than G2G-chain and CUT3R.
  (b) RGB-colored LiDAR points stitched using our predicted poses visualize the boxed region.
  (c) Five-camera \ours{} achieves lower ATE, GPU memory use, and latency than monocular CUT3R.
  Accuracy and costs are measured in separate configurations (Sec.~\ref{sec:exp-efficiency}).}
  \label{fig:teaser}
  \vspace{-0.2cm}
\end{figure}

\section{Introduction}
\label{sec:intro}

Mobile robots and autonomous vehicles often carry synchronized \emph{multi-camera rigs} for surround awareness. Their complementary views support motion estimation when an individual view degrades, while known calibration relates the cameras within the rig. Using these observations online requires causal odometry that treats each rig jointly and estimates motion from current and past inputs with bounded memory and latency.

Classical multi-camera odometry models the rig as a generalized camera within an optimization pipeline \cite{pless2003using,kazik2012real}, but relies on hand-crafted features and careful per-platform engineering. Dynamic, repetitive, and texture-poor scenes remain difficult. 3D foundation models offer geometric priors learned from broad data \cite{wang2024dust3r,wang2025vggt,keetha2025mapanything}. Recent streaming models produce poses causally from monocular video \cite{wang2025cut3r,zhuo2025streamvggt,lan2025stream3r,chen2026gct,cheng2026horizonstream}. Extending this capability to calibrated multi-camera rigs presents two challenges.

The first is to exploit geometry within each rig while keeping temporal inference affordable. Feeding synchronized views into a monocular streamer as separate frames increases the number of tokens processed at each time step. For methods that cache past tokens, it also enlarges the persistent state. Yet this input representation does not explicitly encode the known geometric relations among cameras observing at the same instant. A rig-level streamer must use those relations jointly without carrying all image tokens through its temporal history.

The second is to learn reliable motion estimation from limited rig data. Training a dedicated model from scratch requires both geometric perception and temporal reasoning, but real rig datasets typically provide sparse LiDAR depth, with dense geometric supervision largely confined to simulation. Camera layouts also vary little within each dataset, a limitation noted by Rig3R \cite{li2026rig3r}. Freezing a pretrained front-end removes the need to relearn its 3D representation, but does not by itself teach a new pose back-end to use that representation. Our initialization ablation shows that a randomly initialized back-end learns motion poorly when trained directly on causal rig sequences (Sec.~\ref{sec:exp-ablation}).

Our key insight is to adapt pretrained geometry to streaming through a compact pose back-end. Odometry needs geometric cues that constrain motion, rather than dense reconstruction of every observation, making compact representations suitable for temporal reasoning. Stage~I pretrains this back-end on group relocalization: estimating the relative pose between two groups of views sampled along camera trajectories. Sampling view groups across sessions and trajectories exposes the back-end to diverse geometries and displacements beyond a single fixed rig layout. Stage~II adapts this learned group alignment to causal rig sequences while preserving the frozen front-end's geometric prior.

In this paper, we present \ours{}, a \emph{freeze-and-stream} framework using this insight. To our knowledge, it is the first streaming odometer to transfer a frozen multi-view 3D foundation model to general multi-camera rigs. Its frozen front-end jointly encodes synchronized images with their intrinsics and rig extrinsics. The \emph{Rig-Resampler} compresses each camera's features into a few latent tokens. The \emph{CausalBridge} reasons over this history with a per-query copy of the anchor tokens, an \emph{anchor snapshot}, allowing anchor and query to attend to each other without accessing future observations. Intervening rigs provide geometric context when anchor and query no longer share a view. Poses are estimated relative to the anchor, and the \emph{re-anchoring protocol} periodically advances it to bound the state. Only the compact back-end is trained, using relative pose supervision alone. Mixed-length windows and displacement-normalized translation supervision balance learning across motion ranges.

We evaluate on four rigs: a surround rig on NCLT, a four-way ring on TartanGround, stereo with side fisheyes on KITTI-360, and a humanoid-robot rig. The humanoid experiment tests transfer from simulation to the real robot without training on real-robot data. Figure~\ref{fig:teaser} illustrates the trajectory improvement on NCLT and separately measured runtime costs. Controlled camera-count comparisons and ablations isolate the benefits of joint rig perception, pretraining, anchor interaction, and the training recipe. Replacing the frozen front-end tests the transfer recipe beyond one foundation model. Re-anchoring and training-horizon studies show that longer training windows extend the usable anchor range at greater training cost.

Our contributions are:
\begin{itemize}
\item \textbf{Rig-level causal streaming odometry.} \ours{} combines calibrated multi-view perception with compressed temporal reasoning and re-anchoring for online pose estimation with bounded streaming state.
\item \textbf{A data-efficient transfer recipe.} Group relocalization pretrains a compact pose back-end for causal adaptation on top of a frozen 3D foundation model, using pose supervision alone.
\item \textbf{Accuracy and efficiency across four rigs.} \ours{} achieves the lowest translation and rotation drift among evaluated non-oracle methods on all four rigs; separately measured five-camera inference requires less memory and has lower latency than monocular CUT3R.
\end{itemize}
\begin{figure*}[t]
  \centering
  \includegraphics[width=\textwidth]{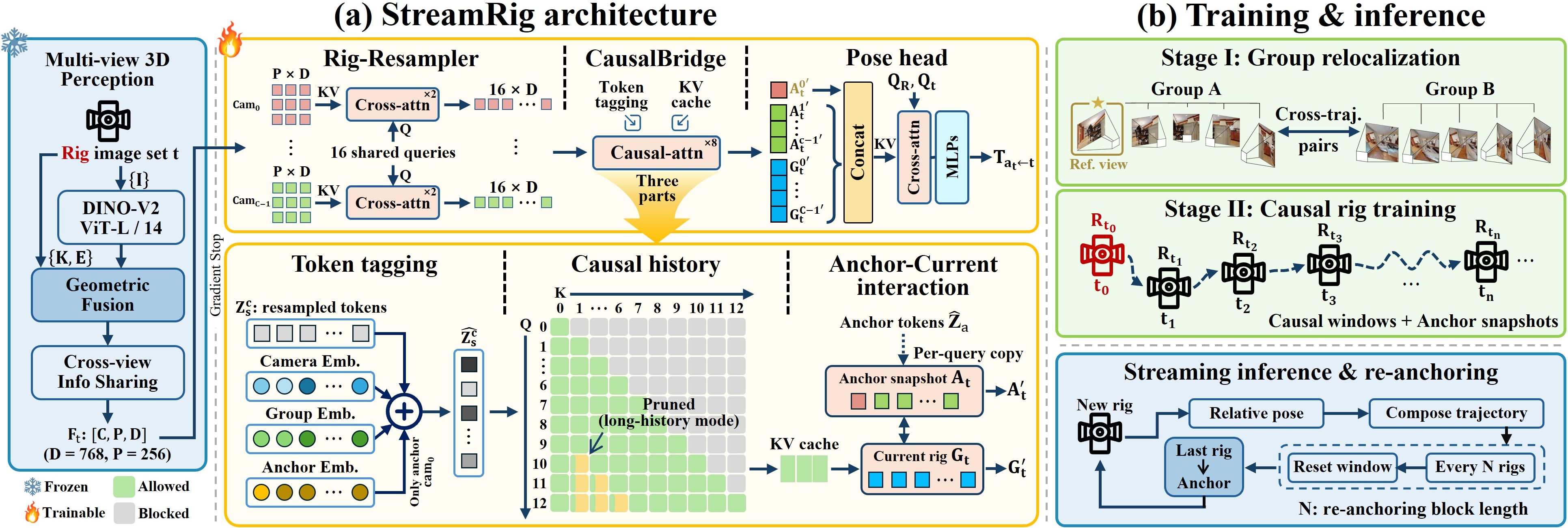}
  \vspace{-0.6cm}
  \caption{\textbf{\ours{} overview.} (a) The frozen front-end perceives each rig with its calibration, the Rig-Resampler compresses every camera into $L$ latent tokens, the CausalBridge attends causally over the compressed rigs, and the pose head regresses $\pose{a}{t}$ from the bridge outputs. Bottom: inside the bridge, tokens are tagged with camera, group, and anchor embeddings, attention across rigs is causal with the keys and values of earlier rigs cached, and every query rig receives its own anchor snapshot; the optional long-history mode prunes the latents of distant rigs. (b) The two training stages, group relocalization with bidirectional attention and causal training on rig windows with the same modules, and the streaming loop that re-anchors every $N$ rigs.}
  \label{fig:arch}
  \vspace{-0.1cm}
\end{figure*}
\section{Related Work}
\label{sec:related}

\textbf{Multi-camera odometry.} Multiple cameras widen the field of view and add redundancy when one view degrades. The generalized camera model \cite{pless2003using} treats a calibrated rig as one imaging device, supporting odometry with little or no view overlap \cite{kazik2012real} and keyframe SLAM with arbitrary rig layouts \cite{kuo2020redesigning,kaveti2023multicamslam}. These systems recover motion by iterative optimization over explicit geometric residuals \cite{campos2021orbslam3}.

\textbf{Feed-forward 3D foundation models} learn geometric priors from large, diverse data to predict geometry and camera parameters in one pass. DUSt3R \cite{wang2024dust3r} predicts dense pointmaps for an uncalibrated image pair, and Reloc3R \cite{dong2025reloc3r} estimates relative poses. VGGT \cite{wang2025vggt} and $\pi^3$ \cite{wang2025pi3} jointly process image collections, the latter without a reference view. Beyond RGB-only inputs, MapAnything \cite{keetha2025mapanything}, Depth Anything 3 \cite{da3team2025depthanything3}, and the Pi3X release of $\pi^3$ accept known camera geometry. Rig3R \cite{li2026rig3r} explicitly conditions on rig metadata: camera identity, timestamps, and rig-relative rays. It demonstrates the benefit of rig priors but identifies limited diversity in training data, particularly in rig configurations, as a key limitation. The group-to-group formulation of \cite{wei2026g2g} freezes the backbone and trains lightweight modules to estimate the rigid transform between two image groups with known intra-group geometry. This enables joint rig perception conditioned on known intrinsics and extrinsics. These methods process fixed image sets without persistent state across calls.

\textbf{Streaming monocular 3D models} process video causally with temporal state. Spann3R \cite{wang2025spann3r} uses external spatial memory queried by each new frame and pruned by attention weight, while CUT3R \cite{wang2025cut3r} maintains a fixed-size recurrent state. TTT3R \cite{chen2026ttt3r} reweights state updates using confidence gating without retraining. StreamVGGT \cite{zhuo2025streamvggt} and STream3R \cite{lan2025stream3r} use causal attention over cached keys and values. WinT3R \cite{li2025wint3r} combines overlapping windows with one camera token per past frame. LingBot-Map \cite{chen2026gct} retains full anchor frames and compresses older frames into summary tokens. HorizonStream \cite{cheng2026horizonstream} combines a decaying recurrent state with a local window. Each processes one image per step and predicts camera parameters instead of using known calibration.

Streaming models thus retain history for single-camera inputs, while rig-aware models jointly perceive calibrated views without temporal state. To our knowledge, prior work has not adapted a frozen multi-view 3D foundation model to causal odometry with calibrated multi-camera rigs. \ours{} fills this gap.
\section{Method}
\label{sec:method}

\ours{} is a causal streaming odometer for calibrated multi-camera rigs. It keeps a multi-view 3D foundation model frozen as the perception front-end and adds three lightweight modules that turn the per-rig perception into a causal state and read odometry out of it (Fig.~\ref{fig:arch}). Only these modules are trained, with relative poses as their sole supervision.

\textbf{Problem formulation.} A \emph{rig} is the set of images captured synchronously by the $C$ cameras of a platform at step $t$, $\mathcal{R}_t = \{(\mathbf{I}_t^c, \mathbf{K}^c)\}_{c=0}^{C-1}$, each with its intrinsics $\mathbf{K}^c$. The extrinsics $\mathbf{E} = \{\pose{0}{c}\}_{c=1}^{C-1}$ relative to a reference camera $0$ are known but not noise-free. Rigs arrive as a causal stream. \ours{} keeps an \emph{anchor} rig $a_t \le t$ as the current reference and realizes the mapping
\begin{equation}
  \big(\mathcal{R}_{a_t}, \dots, \mathcal{R}_t,\, \mathbf{E}\big) \longmapsto
  \pose{a_t}{t} \in \SE,
  \label{eq:mapping}
\end{equation}
the pose of the reference camera of rig $t$ in the frame of the reference camera of the anchor, in one feed-forward pass. The anchor advances with the stream (Sec.~\ref{sec:method-inference}).

\subsection{Rig Perception and Rig-Resampler}
\label{sec:method-perception}

\textbf{Frozen rig perception.} Each incoming rig is encoded once by MapAnything \cite{keetha2025mapanything}, which receives the $C$ images, per-pixel ray directions computed from $\mathbf{K}$, and the extrinsics $\mathbf{E}$ with the reference camera at identity. Inside, a DINOv2 encoder \cite{oquab2024dinov2} tokenizes each image, the ray and pose encodings are fused into the tokens, and an alternating-attention transformer \cite{wang2025vggt} exchanges information across the cameras. The output is a set of geometry-grounded patch features $\mathbf{F}_t \in \R^{C \times P \times D}$, with $P$ patches per camera.

Each rig thus enters the stream as one geometry-grounded observation. Calibration enters \ours{} only through this joint rig perception (Sec.~\ref{sec:exp-rig}); the trainable modules never receive $\mathbf{E}$ or assume a camera layout. Frozen features are cached offline for training. Any multi-view front-end accepting images, intrinsics, and relative poses can supply these features; Sec.~\ref{sec:exp-ablation} tests two alternatives.

\textbf{Rig-Resampler.} Before temporal reasoning, the Rig-Resampler compresses each camera's $P$ patch tokens into $L = 16$ latents using $L$ learnable queries and two cross-attention layers, giving $\mathbf{Z}_t \in \R^{C \times L \times D}$. The queries are shared across cameras, so adding a camera appends $L$ tokens without changing the parameters. A five-camera rig is reduced from 1280 to 80 tokens, fewer than the 256 patches of one $224^2$ image, reducing temporal attention cost and cache size. Such compression targets the cues needed for motion and scale estimation rather than dense reconstruction. Sec.~\ref{sec:exp-ablation} evaluates the trade-off between accuracy and cost at 8, 16, and 32 latents per camera.

\subsection{CausalBridge and Pose Head}
\label{sec:method-bridge}

The CausalBridge holds the causal state of \ours{}: the compressed tokens of the rigs from the anchor $a$ to the current step $t$. Before entering the bridge, the tokens of rig $s$ are tagged with their camera and with the role of the rig,
\begin{equation}
  \hat{\mathbf{Z}}_s^{\,c} \;=\; \mathbf{Z}_s^{\,c}
  \;+\; \mathbf{e}_\text{cam}^{\,c}
  \;+\; \mathbf{e}_\text{group}^{\,\mathbb{1}[s = a]}
  \;+\; \mathbb{1}[s{=}a,\, c{=}0]\,\mathbf{e}_\text{anchor},
  \label{eq:token}
\end{equation}
where the camera embedding is shared by all rigs, the group embedding marks rig $s$ as anchor or query, and the anchor embedding singles out the reference camera of the anchor. The tagged tokens of the window form one sequence, processed by eight self-attention layers under the mask defined next.

\textbf{Causal attention with anchor snapshots.} Attention inside a rig is unrestricted (Fig.~\ref{fig:arch}(a), bottom). Across rigs it is causal: rig $s$ attends to the rigs up to $s$ and never to a later one. A relative pose is read out best when anchor and query attend to each other, but a causal mask never lets the anchor see a later rig. \ours{} therefore gives every query rig $t$ its own copy of the anchor tokens $\hat{\mathbf{Z}}_a$, the \emph{anchor snapshot} $\mathbf{A}_t$. The snapshot attends to rig $t$ alone, and rig $t$ attends to its snapshot and to the rigs after the anchor up to itself,
\begin{equation}
\begin{split}
  \mathcal{V}(\mathbf{A}_t) &= \{\mathbf{A}_t,\, \hat{\mathbf{Z}}_t\},\\
  \mathcal{V}(\hat{\mathbf{Z}}_t) &= \{\mathbf{A}_t\} \cup \{\hat{\mathbf{Z}}_s : a < s \le t\},
\end{split}
  \label{eq:visibility}
\end{equation}
where $\mathcal{V}(\cdot)$ is the set a token group may attend to. The output at step $t$ thus depends only on $\mathcal{R}_a, \dots, \mathcal{R}_t$, so training processes a whole window in one pass and streaming inference adds one rig at a time (Sec.~\ref{sec:method-inference}). Sec.~\ref{sec:exp-ablation} compares this interaction with a static anchor.

\textbf{No temporal position code.} Eq.~\eqref{eq:token} carries no temporal position embedding. Order is conveyed by the causal mask and by the token content. The bridge thus accepts windows with more rigs than seen in training, although the reachable anchor distance stays bounded by the training displacements. Sec.~\ref{sec:exp-ablation} tests adding rotary temporal embeddings.

\textbf{Pose head.} For each camera $c$ of the query rig $t$, two learnable queries, one for rotation and one for translation, cross-attend to the bridge outputs of the reference-camera tokens of the anchor snapshot $\mathbf{A}_t$ and of the tokens of camera $c$, concatenated. Two three-layer MLPs then decode a 6D rotation \cite{zhou2019continuity}, orthogonalized by Gram-Schmidt, and a 3D translation. The odometry output $\pose{a_t}{t}$ is the pose of the reference camera. The other cameras are auxiliary targets whose poses follow from the odometry pose and the known calibration, so they cost no labels and keep the tokens of the whole rig geometrically consistent.

\subsection{Training}
\label{sec:method-training}

\ours{} is supervised by relative poses only. For targets at query offset $d$ with ground truth $(\mathbf{R}, \mathbf{t})$ and prediction $(\hat{\mathbf{R}}, \hat{\mathbf{t}})$, the loss is
\begin{equation}
  \mathcal{L}_d =
  \lambda_R \sqrt{\tfrac{1}{9}\,\mathbb{E}\big\|\hat{\mathbf{R}}^{\top}\mathbf{R} - \mathbf{I}\big\|_F^2}
  + \lambda_t \,
  \mathbb{E}\!\left[\frac{\tfrac{1}{3}\big\|\hat{\mathbf{t}}-\mathbf{t}\big\|_1}
       {\max(\|\mathbf{t}\|,\,\epsilon)}\right],
  \label{eq:loss}
\end{equation}
where $\mathbb{E}$ averages over samples and cameras at fixed $d$; training averages $\mathcal{L}_d$ over $d$. We use $\lambda_R{=}5$, $\lambda_t{=}1$, and $\epsilon{=}0.1$\,m. The rotation term is an RMS chordal loss that approximates a scaled RMS angular error for small residuals. Ground-truth displacement normalization balances near and far translation targets while retaining metric-scale supervision. Sec.~\ref{sec:exp-ablation} tests square-root normalization, no normalization, and up-to-scale supervision.

\textbf{Two-stage training.} To avoid poor convergence from scratch, Stage~I pretrains the three modules on group relocalization with bidirectional attention (Fig.~\ref{fig:arch}(b)). A group is a short window of frames along one camera trajectory, and two groups are paired whenever they share a view, across sessions and trajectories. This exposes the modules to geometries and displacements beyond a single rig layout. Stage~II continues from these weights on the deployed rig, with the modules and their parameters unchanged. Only the input and the supervision change: rig windows with anchor snapshots replace the pair, the causal mask replaces bidirectional attention, and every query rig becomes a target. Throughout Stage~I the input extrinsics are perturbed by $\SE$ noise of 1.5$^\circ$ in rotation and 0.1\,m in translation while all targets stay clean, so the modules learn to recover the true rig geometry from noisy calibration. Stage~II trains on clean extrinsics. Sec.~\ref{sec:exp-ablation} tests omitting Stage~I.

\textbf{Mixed window and stride sampling.} We draw the window length $N$ from $[6,48]$ with probability proportional to $N$, preserving long-range coverage while giving short-range targets greater relative weight. Like displacement normalization, this balances learning across motion ranges. Sec.~\ref{sec:exp-ablation} compares it with fixed $N{=}48$, and Sec.~\ref{sec:exp-reanchor} varies the training horizon. The frame stride is drawn uniformly from $[1, s_{\max}]$, with $s_{\max}$ set per platform by its frame rate and speed.

\textbf{Budget.} The trainable modules hold 74.6\,M parameters, 11.7\% of the 638\,M-parameter system. Stage~II trains on 576\,k windows in total at a total batch size of 16, with AdamW at learning rate $10^{-4}$ and cosine decay. With the frozen features cached, the NCLT model costs about 48 GPU-hours on RTX 4090 GPUs with 24\,GB.

\subsection{Streaming Inference and Re-anchoring}
\label{sec:method-inference}

At each step the backbone perceives the new rig once, the Rig-Resampler compresses it, and its tagged tokens join the active window. The keys and values of every earlier rig are cached in all eight layers of the bridge, so one pass of the bridge over the new rig and its anchor snapshot, followed by the pose head, returns $\pose{a_t}{t}$ (Fig.~\ref{fig:arch}(b), bottom).

\textbf{Re-anchoring protocol.} With a fixed anchor, the state would grow with the sequence. To keep it bounded, \ours{} re-anchors in blocks of $N$ rigs: the last rig of a block becomes the anchor of the next, the window restarts there, and the trajectory is composed from the anchor-to-anchor poses, each of which is one output of Eq.~\eqref{eq:mapping}. The state then holds at most $N$ compressed rigs, or $N\,C L$ tokens, whatever the length of the sequence. Since the bridge accepts any window length, the block length $N$ is a deployment parameter and need not match a training window. Short blocks accumulate more hop errors; long blocks can exceed the learned translation range. Sec.~\ref{sec:exp-reanchor} evaluates this trade-off.

\textbf{Optional long-history mode.} The anchor can also stay fixed over a long stretch, which keeps the whole history addressable at the same per-step latency as re-anchoring. A rig more than 8 steps behind the current one keeps only the first four latents of every camera, the cache drops the keys and values of the rest, and training uses the matching banded mask. The state still grows with the sequence and the pruning costs short-range accuracy, so re-anchoring remains the default and this mode serves as a budget reference (Secs.~\ref{sec:exp-reanchor} and \ref{sec:exp-efficiency}).

\section{Experiments}
\label{sec:exp}

\begin{table*}[t]
  \centering
  \caption{STREAMING ODOMETRY ON FOUR RIGS ($t_{rel}$ IN \%, $r_{rel}$ IN $^\circ$/100\,M, ATE IN M).}
  \label{tab:main}
  \setlength{\tabcolsep}{2.6pt}
  \resizebox{\textwidth}{!}{%
  \begin{tabular}{p{2.9cm}>{\centering\arraybackslash}p{0.85cm}>{\centering\arraybackslash}p{0.85cm}*{12}{>{\centering\arraybackslash}p{0.90cm}}}
  \toprule
  & & & \multicolumn{3}{c}{\makebox[0pt][c]{\textbf{NCLT}}} & \multicolumn{3}{c}{\makebox[0pt][c]{\textbf{TartanGround}}} & \multicolumn{3}{c}{\makebox[0pt][c]{\textbf{KITTI-360}}} & \multicolumn{3}{c}{\makebox[0pt][c]{\textbf{ZJH (real robot)$^{\ddagger}$}}} \\
  & & & \multicolumn{3}{c}{\makebox[0pt][c]{\scriptsize 5-cam surround, co-centered}} & \multicolumn{3}{c}{\makebox[0pt][c]{\scriptsize 4$\times$90$^\circ$ co-located ring}} & \multicolumn{3}{c}{\makebox[0pt][c]{\scriptsize stereo 0.59\,m $+$ 2 fisheye}} & \multicolumn{3}{c}{\makebox[0pt][c]{\scriptsize stereo 7\,cm $+$ 2 oblique}} \\
  & & & \multicolumn{3}{c}{\makebox[0pt][c]{\scriptsize 2 traj. / 12.29 km}} & \multicolumn{3}{c}{\makebox[0pt][c]{\scriptsize 10 scenes / 12.60 km}} & \multicolumn{3}{c}{\makebox[0pt][c]{\scriptsize 2 scenes / 13.63 km}} & \multicolumn{3}{c}{\makebox[0pt][c]{\scriptsize 3 scenes / 0.36 km}} \\
  \cmidrule(lr){4-6} \cmidrule(lr){7-9} \cmidrule(lr){10-12} \cmidrule(lr){13-15}
  Method & In. & Tr. & $t_{rel}\!\downarrow$ & $r_{rel}\!\downarrow$ & ATE$\downarrow$ & $t_{rel}\!\downarrow$ & $r_{rel}\!\downarrow$ & ATE$\downarrow$ & $t_{rel}\!\downarrow$ & $r_{rel}\!\downarrow$ & ATE$\downarrow$ & $t_{rel}\!\downarrow$ & $r_{rel}\!\downarrow$ & ATE$\downarrow$ \\
  \midrule
  \multicolumn{15}{l}{\scriptsize\itshape Non-streaming (offline / pairwise)} \\
  Reloc3R-rig$^{\dagger}$ \cite{dong2025reloc3r} & full & M & 24.26 & 11.06 & 207.4 & 36.39 & 79.18 & 28.6 & 47.12 & 15.02 & 345.4 & 16.49 & 163.17 & 3.77 \\
  Rig3R$^{\dagger}$ \cite{li2026rig3r} & full & M & 48.80 & 24.31 & 216.7 & 158.89 & 231.79 & 65.4 & 471.96 & 34.78 & 4257.0 & 28.02 & 409.79 & 5.23 \\
  MA-A \cite{keetha2025mapanything} & full & U & 46.10 & 28.76 & 227.3 & 29.47 & 90.42 & 24.8 & 11.92 & 4.07 & 169.5 & 11.91 & 75.93 & 2.92 \\
  \rowcolor{gray!8}
  MA-AB~{\scriptsize(oracle)}$^{\mathrm{O}}$ & full & M$^{\mathrm{O}}$ & 45.82 & 27.46 & 331.1 & 2.51 & 5.66 & 3.3 & 4.13 & 2.14 & 56.5 & 4.82 & 45.81 & 1.67 \\
  \prior{}-chain$^{\dagger}$ \cite{wei2026g2g} & full & M & \second{6.08} & \second{3.54} & \second{106.3} & 16.40 & 37.16 & 21.9 & 4.59 & 2.77 & 217.8 & 6.63 & 48.50 & 1.98 \\
  \midrule
  \multicolumn{15}{l}{\scriptsize\itshape Streaming (monocular, front camera)} \\
  CUT3R$^{\dagger}$ \cite{wang2025cut3r} & 1 & M & 9.51 & 4.80 & 241.2 & 22.72 & 29.06 & 21.4 & \second{3.59} & 1.63 & 78.9 & \second{4.61} & \second{39.85} & \second{1.11} \\
  STream3R$^{\dagger}$ \cite{lan2025stream3r} & 1 & U & 20.76 & 4.34 & 171.0 & \second{16.21} & \second{25.56} & \second{13.1} & 5.32 & \second{1.46} & \best{53.6} & 17.02 & 113.72 & 3.78 \\
  LingBot-Map$^{\S}$ \cite{chen2026gct} & 1 & U & 43.77 & 22.86 & 201.4 & 41.61 & 119.37 & 26.4 & 34.60 & 12.00 & 222.1 & 8.26 & 50.23 & 1.73 \\
  HorizonStream \cite{cheng2026horizonstream} & 1 & U & 29.05 & 10.43 & 183.7 & 23.73 & 54.61 & 20.7 & 16.81 & 5.37 & 195.0 & 6.97 & 45.39 & 1.60 \\
  \midrule
  \textbf{\ours{}~(Ours)}$^{\dagger}$ & full & M & \best{2.77} & \best{1.39} & \best{28.4} & \best{12.70} & \best{17.90} & \best{11.0} & \best{2.59} & \best{0.98} & \second{63.7} & \best{3.48} & \best{14.88} & \best{0.88} \\
  \bottomrule
  \end{tabular}%
  }
  \par\vspace{4pt}
  \parbox{\textwidth}{\footnotesize\raggedright
  In.: input cameras per step (full = calibrated rig). Tr.: M = raw metric scale, ATE after SE(3) alignment; U = one global scale per sequence, ATE after Sim(3) alignment; O = ground-truth oracle, excluded from ranking.
  $^{\ddagger}$\,ZJH: every method is evaluated up-to-scale at one shared test stride.
  $^{\dagger}$\,trained on the target training split; $^{\S}$\,released training mixture includes TartanGround and KITTI-360.
  Best in blue and second best in orange.
  }
\end{table*}

\subsection{Setup}
\label{sec:setup}

\textbf{Rigs.} We evaluate on four rigs. \textbf{NCLT} \cite{carlevaris2016university} is a campus Segway with five co-centered surround cameras. We post-train on its eight training sessions and evaluate on the complete sessions 2012-02-19 and 2012-08-20. \textbf{TartanGround} \cite{patel2025tartanground} is a simulated ground robot with four co-located outward cameras. We evaluate the three longest trajectories of each of ten held-out scenes. \textbf{KITTI-360} \cite{liao2021kitti360} is urban driving with a 0.59\,m stereo pair and two side fisheyes rectified to 90$^\circ$ views. We train on drives 0000 and 0002--0007 and evaluate on 0009 and 0010. \textbf{ZJH} is our humanoid-robot dataset for sim-to-real transfer. Its rig has a 7\,cm forward stereo pair and two oblique cameras covering about 180$^\circ$ in azimuth. Training uses only simulation, 131 walking trajectories in 25 indoor environments reconstructed from real scans with 3D Gaussian Splatting. Evaluation is zero-shot on three real sequences in unseen environments, with ground truth from a COLMAP rig reconstruction.

\textbf{Protocol.} Every method runs causally over each complete sequence, anchored to its first pose, and composes its trajectory from its own predictions. Except for the MA-AB oracle, ground-truth poses are used only for evaluation. Each method re-anchors in blocks of $N$ sampled rigs or frames, so $N{=}2$ is a pairwise chain and a larger $N$ keeps a longer state.

\textbf{Metrics.} We report KITTI relative drift, $t_{rel}$ in \% and $r_{rel}$ in $^\circ$ per 100\,m, over segments of 100--800\,m on NCLT and KITTI-360 and 10--80\,m on TartanGround and ZJH, and full-sequence ATE in meters. The two KITTI-360 drives differ three-fold in length, so its metrics are pooled over both weighted by path length. Methods marked U in Table~\ref{tab:main} are not trained on the target dataset or do not output metric scale. They receive one global Umeyama scale per sequence before $t_{rel}$ and $r_{rel}$ are computed, and their ATE follows Sim(3) alignment. All others run at raw metric scale, with ATE after SE(3) alignment. ZJH is the exception, since the scale between simulation and the real robot is unknown in advance, so there every method receives one global scale per sequence and shares one test stride.

\textbf{Baselines.} \emph{Streaming, monocular:} CUT3R \cite{wang2025cut3r} and STream3R \cite{lan2025stream3r}, fine-tuned on each target split with their original recipes and native nine-frame window, and LingBot-Map \cite{chen2026gct} and HorizonStream \cite{cheng2026horizonstream} with released weights. \emph{Non-streaming, rig-aware:} Reloc3R \cite{dong2025reloc3r} and Rig3R \cite{li2026rig3r} trained on the target split; MapAnything \cite{keetha2025mapanything} as MA-A, given the images and intrinsics of both rigs but the extrinsics of the anchor rig only, and as MA-AB, an oracle also given the ground-truth extrinsics of the query rig and hence the relative pose itself; and \prior{}-chain, which chains the pairwise estimator of \prior{} \cite{wei2026g2g} rig to rig.

\subsection{Main Results}
\label{sec:exp-main}
\label{sec:exp-rig}

\begin{table}[t]
  \centering
  \caption{MONO, STEREO, AND FULL RIG ON KITTI-360.}
  \label{tab:rig}
  \setlength{\tabcolsep}{3pt}
  \small
  \renewcommand{\arraystretch}{1.15}
  \begin{tabular}{@{}p{0.36\columnwidth} *{4}{>{\centering\arraybackslash}p{\dimexpr0.16\columnwidth-6pt\relax}}@{}}
  \toprule
  Cameras & $t_{rel}\!\downarrow$ & $r_{rel}\!\downarrow$ & SE3$\downarrow$ & Sim3$\downarrow$ \\
  \midrule
  Mono & 4.84 & 1.92 & 85.4 & 78.5 \\
  Stereo & 3.59 & 1.33 & 67.2 & 65.6 \\
  \textbf{Stereo $\boldsymbol{+}$ 2 side} & \textbf{2.59} & \textbf{0.98} & \textbf{63.7} & \textbf{59.1} \\
  \bottomrule
  \end{tabular}
  \par\vspace{4pt}
  \parbox{\columnwidth}{\footnotesize\raggedright
  KITTI-360 row of Table~\ref{tab:main}, changing only the camera subset entering the frozen front-end; full-drive ATE after SE(3) / Sim(3) alignment.
  }
\end{table}

\begin{figure*}[t]
  \centering
  \includegraphics[width=0.9\textwidth]{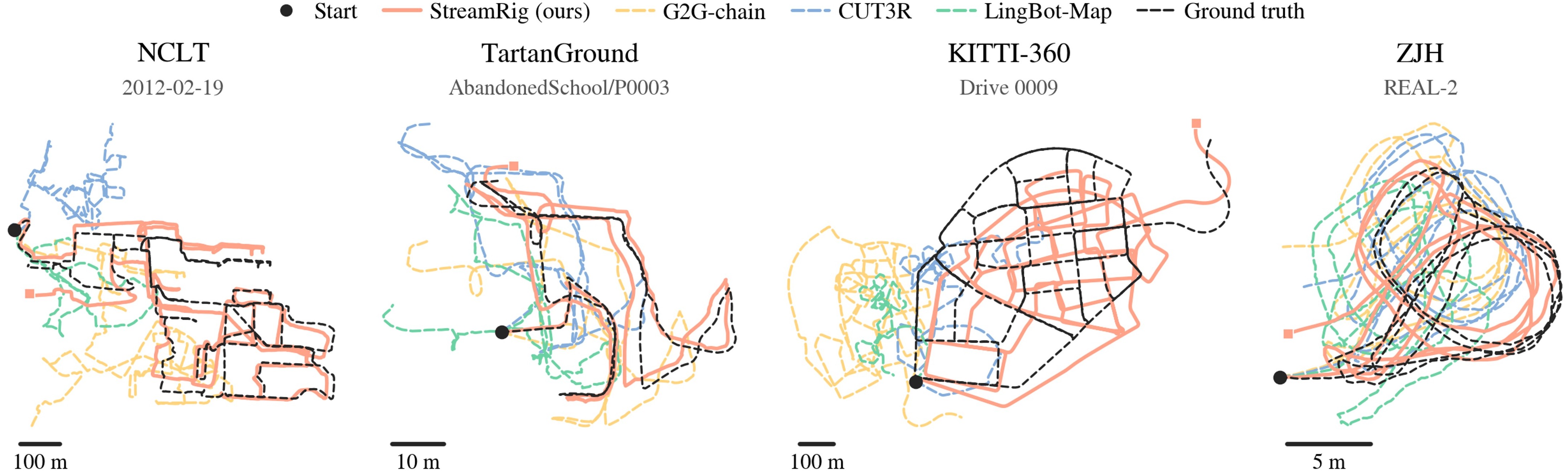}
  \vspace{-0.2cm}
  \caption{Example trajectories from the Table~\ref{tab:main} models, after full-sequence Sim(3) alignment and translation to a common start; KITTI-360 follows the same GT gap-bridging protocol as Table~\ref{tab:main}.}
  \label{fig:qualitative-trajectories}
  \vspace{0cm}
\end{figure*}

\begin{figure}[t]
  \centering
  \includegraphics[width=\columnwidth]{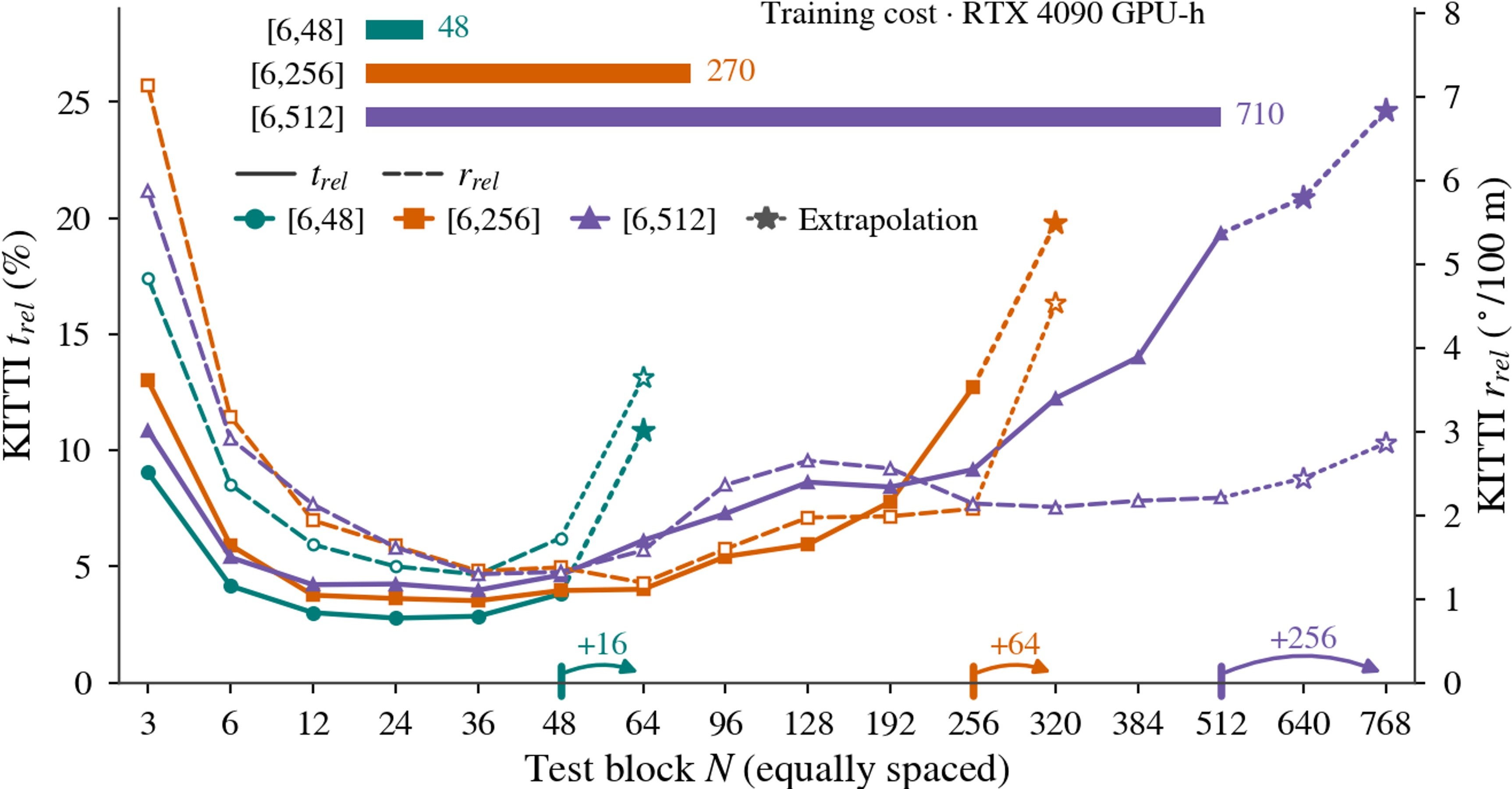}
  \vspace{-0.6cm}
  \caption{\textbf{Re-anchoring interval and training horizon on NCLT.} Translation drift (filled, left axis) and rotation drift (open, right axis) versus re-anchoring interval. Stars mark extrapolation beyond training limits. Top bars show estimated training costs in RTX~4090 GPU-hours.}
  \label{fig:reanchor}
  \vspace{-0.2cm}
\end{figure}

Table~\ref{tab:main} compares joint rig perception and temporal state: rig-aware pairwise methods provide the former, monocular streamers the latter. \ours{} combines both and achieves the lowest translation and rotation drift among non-oracle methods on all four rigs.
Figure~\ref{fig:qualitative-trajectories} illustrates the recovered trajectory shapes and accumulated drift on all four rigs after scale correction.

\textbf{NCLT} is the hardest rig. Its two 6\,km sessions are real campus captures with indoor-outdoor transitions, lighting changes, motion blur, and pedestrians \cite{carlevaris2016university}. Surround perception of the rig first buys stable tracking here. \prior{}-chain keeps no state over time, yet it is second in the column and drifts less than every monocular streamer, fine-tuned or zero-shot. The streaming state then turns stable tracking into low drift. \ours{} sees the same images, calibration, and frozen features as the chain, but reads each rig against an anchor several rigs back through the causal history in between, so every estimate draws on more evidence and the anchor changes far less often. A chain re-estimates at every hop and composes the results, and composition accumulates error. Even MA-AB, given the true pose between the rigs of every pair, still drifts 46\%, because a small systematic bias per pairwise estimate is composed over thousands of hops.

\textbf{TartanGround} stresses the same two ingredients through motion rather than appearance. The simulated robot climbs slopes and crosses vegetation and clutter where occlusions come and go \cite{patel2025tartanground}, so a wide view of the rig and a memory of what was seen both matter. \prior{}-chain drifts far less than Reloc3R and Rig3R, which are retrained on the target split, so the frozen pre-trained front-end is what carries rig perception to new scenes. The fine-tuned monocular streamers CUT3R and STream3R reach a similar level, which is what a history adds over a single pair. \ours{} combines the two and leads the column.

\textbf{KITTI-360} is forward driving, the steadiest motion of the four rigs, above all in rotation, and monocular methods do well. CUT3R and STream3R take the second-best translation and rotation drift. Table~\ref{tab:rig} isolates the cameras. With the front camera alone our model trails CUT3R, and adding the stereo pair draws level. The second view adds little field of view, but joint perception of two calibrated views carries depth that a single view has to infer. The two side cameras widen the field of view, and the full rig leads among our camera configurations on every metric, including the complete-drive ATE. Across methods in Table~\ref{tab:main}, STream3R reaches the lowest ATE, measured after Sim(3) alignment.

\textbf{ZJH} shows the zero-shot sim-to-real transfer of \ours{}. A walking humanoid is a harder platform than a wheeled one. Every step shakes the rig, and the turn per meter of travel is the largest of the four rigs. On the real humanoid \ours{} has the lowest translation drift, rotation drift, and ATE in the column. Freezing the foundation model lets \ours{} learn rig modeling and odometry in simulation and carries the generalization of the front-end to the real robot. The gain is largest in rotation, where \ours{} lowers the drift of the second-best method, CUT3R, by 63\%. The streaming state matters on this platform. CUT3R is ahead of \prior{}-chain, and the zero-shot streamers LingBot-Map and HorizonStream rank ahead of every non-streaming row except \prior{}-chain, since a history over many rigs absorbs the perturbation the gait adds to every step.

\subsection{Re-anchoring Distance and Training Horizon}
\label{sec:exp-reanchor}

\textbf{Re-anchoring distance.} Fig.~\ref{fig:reanchor} sweeps the re-anchoring block $N$ over the two complete NCLT sessions. All three models show a U-shaped translation-error curve, with their lowest drift at intermediate blocks. Anchors too close make the trajectory a chain of many short hops, each carrying the noise floor of a single estimate. At larger separations, anchor and query may no longer share a field of view, so their relation must be relayed through the rigs in between, and one-shot pose regression becomes harder as displacement grows. The standard configuration, trained on $[6,48]$ rigs, has its lowest translation drift near $N{=}24$; drift rises modestly by $N{=}48$ and sharply at $N{=}64$, beyond its training horizon. Displacement diagnostics support a learned-range limitation: emitted displacements saturate near the largest training displacements as test targets cross that range. The training horizon therefore influences the rising side of the U.

\textbf{Training horizon.} Longer-window training shifts the large-block rise to the right, at additional cost. We extend the training windows to $[6,256]$ and $[6,512]$ rigs. The $[6,256]$ model has a broad low-error region at intermediate blocks, but translation drift already rises within its training range and increases sharply from $N{=}256$ to 320. The $[6,512]$ model is evaluated out to $N{=}768$ and rises more gradually beyond its limit, although its translation drift is already high near $N{=}512$. This gentler rise occurs because its training displacement range already exceeds the separation of the vast majority of frame pairs within individual NCLT trajectories. Longer training thus extends the usable block range without improving the minimum drift or guaranteeing low error throughout the training horizon. At these distances, turns can make anchor-to-query displacement much shorter than the traveled path; the bridge relates the query to the anchor through intervening rigs, while displacement coverage still limits the pose head. This extended range costs training time: the longer models see the same number of rigs per epoch in fewer, longer windows, but use two and three times as many epochs, and hundreds of rigs per window cost more per rig in the long-history mode of Sec.~\ref{sec:method-inference}. Estimated RTX~4090 training costs rise from 48 GPU-hours to 270 and 710.

\subsection{Ablations}
\label{sec:exp-ablation}

\begin{table}[t]
  \begingroup
  \makeatletter
  \renewenvironment{table}[1][]{\def\@captype{table}}{}
  \makeatother
  {\begin{table}[t]
  \centering
  \caption{ABLATIONS, ONE FACTOR AT A TIME.}
  \label{tab:ablation}
  \small
  \renewcommand{\arraystretch}{1.25}
  \setlength{\tabcolsep}{1.5pt}
  \begin{tabular*}{\columnwidth}{@{\extracolsep{\fill}} l l c@{\hspace{4pt}}c c@{\hspace{4pt}}c c@{\hspace{4pt}}c @{}}
  \toprule
   & & \multicolumn{2}{c}{NCLT} & \multicolumn{2}{c}{TartanGround} & \multicolumn{2}{c}{ZJH (real)$^{\S}$} \\
  \cmidrule(lr){3-4} \cmidrule(lr){5-6} \cmidrule(lr){7-8}
  Group & Variant & $t_{rel}$ & $r_{rel}$ & $t_{rel}$ & $r_{rel}$ & $t_{rel}$ & $r_{rel}$ \\
  \midrule
  \multicolumn{2}{l}{\textbf{Ours (full)}} & \textbf{2.77} & 1.39 & \textbf{12.70} & 17.90 & 3.48 & \textbf{14.9} \\
  \midrule
  Window & Fixed $N{=}48$ & 3.33 & 1.89 & 14.58 & 20.65 & 4.54 & 23.1 \\
  \cmidrule(lr){1-8}
  Anchor & Static & 3.41 & 2.02 & 13.83 & 18.26 & 4.79 & 27.9 \\
  \cmidrule(lr){1-8}
  \multirow{3}{*}{$t$ loss} & Up-to-scale$^{\ddagger}$ & 6.87 & 2.26 & 27.51 & 15.99 & 4.49 & 28.2 \\
   & Unnormalized $t$ & 3.55 & 1.75 & 14.26 & 25.69 & 5.46 & 24.7 \\
   & $\sqrt{\lVert t\rVert}$ norm. & 3.32 & 1.89 & 13.89 & 18.71 & 4.47 & 22.6 \\
  \cmidrule(lr){1-8}
  \multirow{2}{*}{Latents} & 8 / cam & 3.27 & 1.53 & 15.03 & 19.47 & 3.66 & 18.4 \\
   & 32 / cam & 2.82 & \textbf{1.08} & 13.51 & \textbf{13.30} & \textbf{3.16} & 18.8 \\
  \cmidrule(lr){1-8}
  Pos.\ code & RoPE & 2.98 & 1.62 & 13.01 & 18.05 & 3.65 & 22.9 \\
  \cmidrule(lr){1-8}
  Init & No warm-start & 41.37 & 22.64 & 57.99 & 237.59 & 16.80 & 102.6 \\
  \bottomrule
  \end{tabular*}
  \par\vspace{4pt}
  \parbox{\columnwidth}{\footnotesize\raggedright
  $^{\ddagger}$\,scored with one global scale per sequence.
  $^{\S}$\,zero-shot on the real robot, scored with one global scale per sequence.
  }
\end{table}}
  \par\vspace{-4pt}
  {\begin{table}[t]
  \centering
  \caption{FROZEN FRONT-END ON NCLT.}
  \label{tab:scaling}
  \setlength{\tabcolsep}{2.4pt}
  \scriptsize
  \resizebox{\columnwidth}{!}{%
  \begin{tabular}{l cccc cc}
  \toprule
  & \multicolumn{4}{c}{$t_{rel}$ (\%) at $N$} & \multicolumn{2}{c}{Best} \\
  \cmidrule(lr){2-5} \cmidrule(lr){6-7}
  Variant & 6 & 12 & 24 & 48 & $t_{rel}$ & $r_{rel}$ \\
  \midrule
  ~~\textbf{MapAnything (default)} & \textbf{5.32} & \textbf{4.97} & \textbf{4.45} & \textbf{6.33} & \textbf{4.20} & \textbf{2.18} \\
  ~~Depth Anything 3 & 7.15 & 5.91 & 5.62 & 7.94 & 5.61 & 2.76 \\
  ~~$\pi^3$X & 9.67 & 7.28 & 6.99 & 9.87 & 6.78 & 3.35 \\
  \bottomrule
  \end{tabular}%
  }
  \par\vspace{4pt}
  \parbox{\columnwidth}{\footnotesize\raggedright
  Best: minimum tested $t_{rel}$ with paired $r_{rel}$. HM3D-Reloc Stage~I initialization; 10 training epochs (20 for $\pi^3$X).
  }
\end{table}
}
  \par\vspace{-8pt}
  \endgroup
\end{table}

Table~\ref{tab:ablation} changes one factor at a time on NCLT, TartanGround and ZJH, and Table~\ref{tab:scaling} swaps the frozen front-end. The ZJH column keeps the zero-shot real-robot protocol of Table~\ref{tab:main}.

\textbf{Window curriculum and translation supervision.} A far query is related to the anchor through the rigs in between, so an accurate long-range estimate rests on accurate short-range relations in the cached state. The window sampling and translation loss of Sec.~\ref{sec:method-training} control how much gradient the short range receives. A fixed window of 48 rigs weights every target from 1 to 47 hops equally, so the far targets, whose errors are naturally larger, take over the loss. An unnormalized translation loss does the same through displacement, since far targets have larger absolute errors. Either way the near range is under-trained, the far targets then read from an inaccurate state, and the error grows at both ends. Mixed windows and displacement normalization counter this imbalance. The fixed window costs 15 to 30\% across the three rigs, a square-root normalization about as much, and no normalization up to twice that. An up-to-scale translation loss in the style of STream3R \cite{lan2025stream3r}, scored with one global scale per sequence like the up-to-scale rows of Table~\ref{tab:main}, has higher translation drift on all three rigs. MapAnything already outputs metric-scale geometry, so \ours{} can be supervised at metric scale, and this supervision improves the trajectory beyond fixing its scale.

\textbf{Anchor snapshot.} Replacing the per-query snapshot of Sec.~\ref{sec:method-bridge} with a static anchor costs 9 to 38\% across the three rigs. A static anchor is read by the query but remains query-independent throughout the bridge. The snapshot lets the two rigs attend to each other through the whole bridge, where the relative pose is resolved.

\textbf{Latents per camera.} Halving the latents from 16 to 8 costs up to 18\% in translation. Doubling them to 32 leaves translation within a few percent and lowers rotation drift by about a quarter on NCLT and TartanGround, at about three times the training compute, so we keep 16 per camera.

\textbf{Temporal position code.} A rotary temporal code brings no gain on any rig, and translation drifts up to 8\% more than with the code-free bridge. A monocular streamer recovers geometry across time from single frames, so it must know which frames lie close in time, and LingBot-Map loses clearly without the rotary code in its trajectory memory \cite{chen2026gct}. \ours{} resolves geometry inside each rig with the frozen front-end, and the anchor snapshot names the reference rig explicitly, so the bridge only relates rig-level observations and needs no order beyond the causal mask.

\textbf{Initialization.} Removing Stage~I increases translation drift by approximately 4.6--14.9$\times$ under the same training setup. Frozen geometric features alone therefore do not yield an effective pose readout when trained from scratch. Group relocalization on diverse layouts and displacements teaches the back-end to align geometric features before causal adaptation to a fixed rig (Sec.~\ref{sec:method-training}).

\textbf{Frozen front-end.} Table~\ref{tab:scaling} tests the front-end interface of Sec.~\ref{sec:method-perception} by substituting Depth Anything 3 \cite{da3team2025depthanything3} or $\pi^3$X \cite{wang2025pi3}. Both alternatives converge, stream, and trace the same U over $N$. The three front-ends differ in how directly they expose intra-rig geometry to the readout, with per-pixel ray conditioning the most direct, and we read their ordering as a statement about that rather than about model size.

\subsection{Compute and Memory Budget}
\label{sec:exp-efficiency}

\begin{table}[t]
  \centering
  \caption{COMPUTE AND MEMORY BUDGET PER ARRIVAL.}
  \label{tab:efficiency}
  \setlength{\tabcolsep}{3pt}
  \footnotesize
  \resizebox{\columnwidth}{!}{%
  \begin{tabular}{l c *{3}{>{\centering\arraybackslash}p{0.105\columnwidth}}}
  \toprule
  Method & Input & ms$\downarrow$ & Thr.$\uparrow$ & Mem$\downarrow$ \\
  \midrule
  \multicolumn{5}{l}{\itshape Streaming (monocular)} \\
  CUT3R (recurrent)      & $224{\times}224$        & 31.3  & 31.9 & 3.9 \\
  STream3R (window 5)    & $224{\times}224$        & 57.6  & 17.4 & 6.6 \\
  STream3R (full causal) & $224{\times}224$ & 121.4 & 8.2 & OOM \\
  LingBot-Map            & $518{\times}378$ & 150.5 & 6.6  & 13.8 \\
  HorizonStream          & $504{\times}378$ & 935.3 & 1.1  & 15.0 \\
  \midrule
  \multicolumn{5}{l}{\itshape Five-camera rig} \\
  \prior{}-chain          & $5{\times}224^{2}$ & 50.9 & 19.6 & 2.6 \\
  \textbf{\ours{}} (default)  & $5{\times}224^{2}$ & \textbf{26.2} & \textbf{38.2} & \textbf{2.6} \\
  \ours{} (512-history)                     & $5{\times}224^{2}$ & 26.4 & 37.9 & 3.0 \\
  \bottomrule
  \end{tabular}%
  }
  \par\vspace{4pt}
  \parbox{\columnwidth}{\footnotesize\raggedright
  RTX~4090, batch~1, BF16, 512 arrivals. ms: median arrival latency; Thr.: frames/s (monocular) or rigs/s (five-camera); Mem: peak allocated GiB.
  }
\end{table}

Figure~\ref{fig:teaser} uses NCLT 2012-08-20 for accuracy, with \ours{} at $N=24$ and fine-tuned CUT3R using a nine-frame window. Its costs are measured separately on an RTX 4090 with BF16 and batch size one, using \ours{} at $N=2$ and recurrent CUT3R. The plotted LiDAR points visualize the estimated poses and are not odometry inputs.

Table~\ref{tab:efficiency} measures every method on the same GPU with the same inputs resident in memory, from RGB to pose, over 512 arrivals. \ours{} processes a five-camera rig in 26.2\,ms with 2.6\,GiB, whereas CUT3R, the fastest monocular streamer, spends 31.3\,ms and 3.9\,GiB on one frame. The window-based and long-context streamers cost several times more. Perception dominates. The frozen backbone and the Rig-Resampler take 24.9 of the 26.2\,ms and the CausalBridge with the pose head about 1.3\,ms, so the temporal module is almost free and the marginal cost of a camera is the cost of perceiving it.

A pairwise chain that re-encodes both rigs of every pair costs 1.9 times more, but a chain that caches the features of the previous rig does exactly the work of \ours{} at $N{=}2$, so the gain of Table~\ref{tab:main} is accuracy at equal cost, not speed. The optional long-history mode keeps the whole history addressable at the same per-step latency, but its state grows with the sequence and peaks at 3.0\,GiB after 512 rigs. Bounded memory comes from re-anchoring, not from compression alone.

\section{Conclusion}
\label{sec:conclusion}

The results of \ours{} support joint rig perception and temporal history as complementary sources of odometry accuracy. The Rig-Resampler and CausalBridge make this combination affordable, while group relocalization provides the alignment initialization that limited rig data alone does not supply. The re-anchoring study shows that longer training windows extend the usable anchor range. Effective streaming combines compact geometric observations with an appropriate re-anchoring interval.

\textbf{Future work.} Exploratory trials on NCLT suggest that compressed rig tokens may be useful for same-session revisit retrieval, but we do not quantitatively evaluate this capability here, and acceptance thresholds still require per-session tuning. We will develop transferable loop triggers and integrate revisits into the streaming state. We will also jointly refine poses and dense geometry so that the accumulated map constrains subsequent estimates.

\bibliographystyle{IEEEtran}
\bibliography{references}

\end{document}